\documentclass{article}

\usepackage{graphicx} 
\usepackage{subfigure} 

\usepackage{natbib}

\usepackage{algorithm}
\usepackage{algorithmic}

\usepackage{hyperref}

 \usepackage[accepted]{techreport}

\icmltitlerunning{}

\usepackage{amsmath,amssymb}

\usepackage{color}
\usepackage[all]{xy}
\usepackage{verbatim}
\usepackage{cancel}
\definecolor{Blue}{rgb}{0.9,0.3,0.3}

\newcommand{\squishlist}{
   \begin{list}{$\bullet$}
    { \setlength{\itemsep}{0pt}      \setlength{\parsep}{3pt}
      \setlength{\topsep}{3pt}       \setlength{\partopsep}{0pt}
      \setlength{\leftmargin}{1.5em} \setlength{\labelwidth}{1em}
      \setlength{\labelsep}{0.5em} } }

\newcommand{\squishlisttwo}{
   \begin{list}{$\bullet$}
    { \setlength{\itemsep}{0pt}    \setlength{\parsep}{0pt}
      \setlength{\topsep}{0pt}     \setlength{\partopsep}{0pt}
      \setlength{\leftmargin}{2em} \setlength{\labelwidth}{1.5em}
      \setlength{\labelsep}{0.5em} } }

\newcommand{\squishend}{
    \end{list}  }

\newcommand{\myvec}[1]{\mathbf{#1}}

\newcommand{\vf}{\myvec{f}}

\newcommand{\vk}{\myvec{k}}

\newcommand{\vw}{\myvec{w}}

\newcommand{\vD}{\myvec{D}}

\newcommand{\vK}{\myvec{K}}

\newcommand{\be}{\begin{equation}}
\newcommand{\ee}{\end{equation}}
\newcommand{\bea}{\begin{eqnarray}}
\newcommand{\eea}{\end{eqnarray}}
\newcommand{\beaa}{\begin{eqnarray*}}
\newcommand{\eeaa}{\end{eqnarray*}}

\DeclareMathAlphabet{\mathpzc}{OT1}{pzc}{m}{n}

\newcommand{\fib}[3]{%
 \xy(0,5) \xymatrix@=+10pt{%
 #1 \ar[d]_{#2} \\
 #3             } \endxy }

\newcommand{\mapdef}[5]{%
\xy(0,4) \xymatrix@R=+0pt{%
 \negthickspace\negthickspace\negthickspace    #1 : #2    \ar[r]        &  #3  \\
 \quad #4                                                 \ar@{|->}[r]  &  #5  } \endxy }

\newcommand{\mapdefDisp}[5]{%
\xymatrix@R=+0pt{%
 \negthickspace\negthickspace\negthickspace    #1 : #2    \ar[r]        &  #3  \\
 \quad #4                                                 \ar@{|->}[r]  &  #5  } }

\newcommand{\mc}[1]{\mathcal{#1}}

\newcommand{\bb}[1]{\text{$\mathbb #1$}}
\newcommand{\bs}[1]{\text{$\boldsymbol{#1}$}}

\newcommand{\qqquad}{\qquad\qquad}

\newcommand{\mat}[1]{\begin{bmatrix} #1 \end{bmatrix} }

\usepackage{theorem}
\newcommand{\BlackBox}{\rule{1.5ex}{1.5ex}}  
\newenvironment{proof}{\par\noindent{\bf Proof\ }}{\hfill\BlackBox\\[2mm]}
 
\newtheorem{theorem}{Theorem}
\newtheorem{lemma}[theorem]{Lemma} 
 
\newtheorem{proposition}[theorem]{Proposition} 
\newtheorem{remark}[theorem]{Remark}
\newtheorem{corollary}[theorem]{Corollary}
\newtheorem{definition}[theorem]{Definition}

\newcommand{\inner}[2]{\left\langle #1,#2 \right\rangle}

\newcommand{\cbr}[1]{\left\{#1\right\}}
\newcommand{\nbr}[1]{\left\|#1\right\|}
\newcommand{\abr}[1]{\left|#1\right|}

\newcommand{\one}{\mathbf{1}}

\newcommand{\Bcal}{\mathcal{B}}

\newcommand{\Hcal}{\mathcal{H}}

\DeclareMathOperator*{\argmax}{\mathrm{argmax}}

\newcommand{\intset}[1]{\cbr{1..n}}

\usepackage{geometry}
\usepackage{wrapfig}

\usepackage{color}

\newcommand{\sfrac}[2]{\leavevmode\kern.1em
           \raise.5ex\hbox{\footnotesize #1}\kern-.1em
                   /\kern-.15em\lower.25ex\hbox{\footnotesize #2}}

\def\capstyle#1{\small \emph{#1}}

\DeclareMathOperator{\diam}{diam}
\DeclareMathOperator{\cvx}{conv}
\DeclareMathOperator{\I}{I}
\DeclareMathOperator*{\Span}{span}
\DeclareMathOperator*{\argsup}{argsup}

\newcommand{\oneoftwo}[2]{#1}  

\begin{document}

 \onecolumn
 
\icmltitle{Regret Bounds for Deterministic Gaussian Process Bandits}

\icmlauthor{Nando de Freitas}{nando@cs.ubc.ca}
\icmladdress{Department of Computer Science, University of British Columbia, Vancouver, BC V6T 1Z4, Canada}
\icmlauthor{Alex J. Smola}{alex@smola.org}
\icmladdress{Yahoo! Research, Santa Clara, CA 95051, USA}
\icmlauthor{Masrour Zoghi}{mzoghi@cs.ubc.ca}
\icmladdress{Department of Computer Science, University of British Columbia, Vancouver, BC V6T 1Z4, Canada}

\icmlkeywords{Bayesian optimization, Gaussian Process bandits, global optimization}

\vskip 0.3in
 

\begin{abstract}
This paper analyzes the problem of Gaussian process (GP) bandits with deterministic observations. The analysis uses a branch and bound algorithm that is related to the UCB algorithm of \cite{Srinivas2010gp}. For GPs with Gaussian observation noise, with variance strictly greater than zero, \cite{Srinivas2010gp} proved that the regret vanishes at the approximate rate of
 $\mc O\left(\frac{1}{\sqrt{t}}\right)$, where $t$ is the number of observations. To complement their result, we attack the deterministic case and attain a much faster exponential convergence rate. Under some regularity assumptions,
we show that the regret decreases asymptotically according to $\mc O\left(e^{-\frac{\tau t}{\left(\ln t\right)^{d/4}}}\right)$ with high probability. Here, $d$ is the dimension of the search space and $\tau$ is a constant that depends on the behaviour of the objective function near its global maximum.
\end{abstract}

\section{Introduction}
\label{sec:intro} 


Let $f: \mc D \to \bb R$ be a function on a compact subset $\mc D \subseteq \bb R^d$. We would like to address the global optimization problem
\[ x_M = \argmax_{x \in \mc D} f(x). \]
Let us assume for the sake of simplicity that the objective function $f$ has a unique global maximum (although it may have many local maxima).

The space $\mc D$ might be the set of free parameters that one could feed into a time-consuming algorithm or the locations where a sensor could be deployed, and the function $f$ might be a measure of the performance of the algorithm (e.g. how long it takes to run). We refer the reader to \cite{Mockus-82,Schonlau-98,Gramacy-04,Brochu07,Lizotte-08,Martinez-Cantin-09,Garnett-10} for many practical examples of this global optimization setting. In this paper, our assumption is that once the function has been probed at point $x \in \mc D$, then the value $f(x)$ can be observed with very high precision. This is the case when the deployed sensors are very accurate or if the algorithm is deterministic. An example of this is the configuration of CPLEX parameters in mixed-integer programming \cite{Hutter2010mip}. More ambitiously, we might be interested in the \emph{simultaneous} automatic configuration of an entire system (algorithms, architectures and hardware) whose performance is deterministic in terms of several free parameters and design choices.



Global optimization is a difficult problem without any assumptions on the objective function $f$. The main complicating factor is the uncertainty over the extent of the variations of $f$, e.g. one could consider the characteristic function, which is equal to $1$ at $x_M$ and $0$ elsewhere, and none of the methods we mention here can optimize this function without exhaustively searching through every point in $\mc D$.

The way a large number of global optimization methods address this problem is by imposing some prior assumption on how fast the objective function $f$ can vary. The most explicit manifestation of this remedy is the imposition of a Lipschitz assumption on $f$, which requires the change in the value of $f(x)$, as the point $x$ moves around, to be smaller than a constant multiple of the distance traveled by $x$ \cite{LipschitzReview1992}. As pointed out in \citep[Figure 3]{Xarmed2011}, it is only important to have this kind of tight control over the function near its optimum: elsewhere in the space, we can have what they have dubbed a ``weak Lipschitz'' condition.

One way to relax these hard Lipschitz constraints is by putting a Gaussian Process (GP) prior on the function. Instead of restricting the function from oscillating too fast, a GP prior requires those fast oscillations to have low probability, cf. \citep[Theorem 5]{GhosalRoy2006}. 
\begin{figure*}[t!]
\begin{center}
  \includegraphics[width=\textwidth]{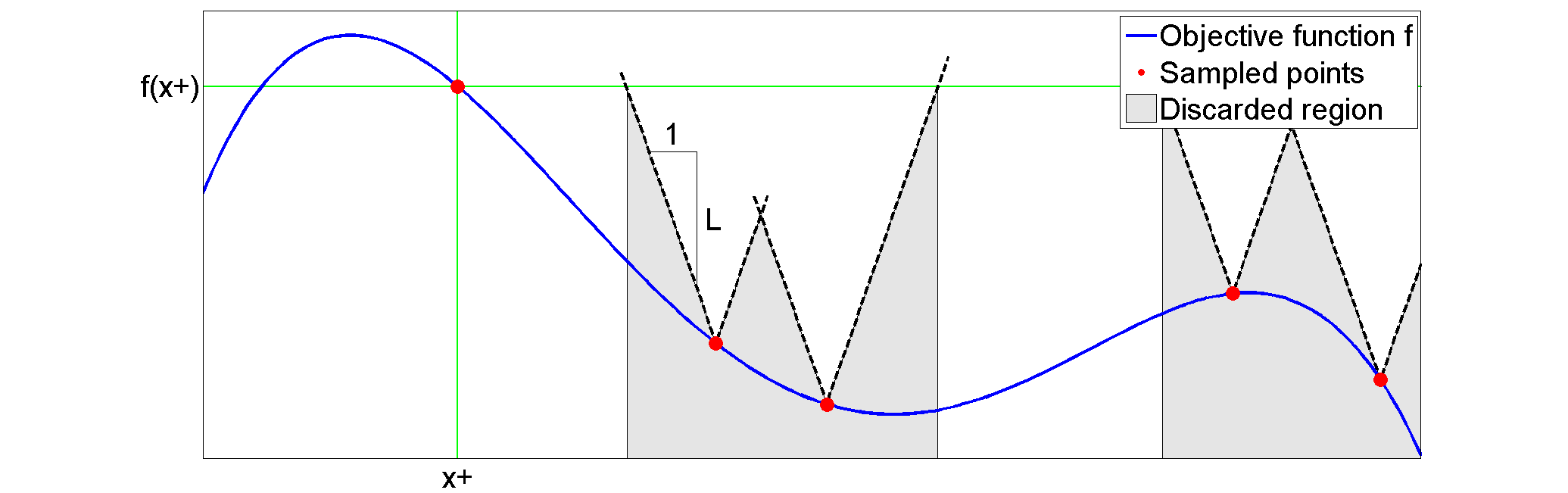}
\end{center}
\caption{\capstyle{An example of the Lipschitz hypothesis being used to discard pieces of the search space when finding the maximum of a function $f$. Although $f$ is only known at the red sample points, if the derivative upper bounds (dashed lines) are below the best attained value thus far, $f(x^{+})$, the corresponding areas of the search space (shaded regions) may be discarded. 
}} 
\label{fig:RelevantSet}
\end{figure*}

The main point of these bounds (be they hard or soft) is to assist with the \emph{exploration-exploitation trade-off} that global optimization algorithms have to grapple with. In the absence of any assumptions of convexity on the objective function, a global optimization algorithm is forced to explore enough until it reaches a point in the process when with some degree of certainty it can localize its search space and perform local optimization (exploitation). Derivative bounds such as the ones discussed here together with the boundedness of the search space, guaranteed by the compactness assumption on $\mc D$, provide us with such certainty by producing a useful upper bound that allows us to shrink the search space. 
This is illustrated in Figure \ref{fig:RelevantSet}. Suppose we know that our function is Lipschitz with constant $L$, then given sample points as shown in the figure, we can use the Lipschitz property to discard pieces of the search space. This is done by finding points in the search space where the function could not possibly be higher than the maximum value already encountered. Such points are found by placing cones at the sampled points with slope equal to $L$ and checking where those cones lie below the maximum observed value.

\begin{figure*}[t]
\begin{center}
  \includegraphics[width=\textwidth]{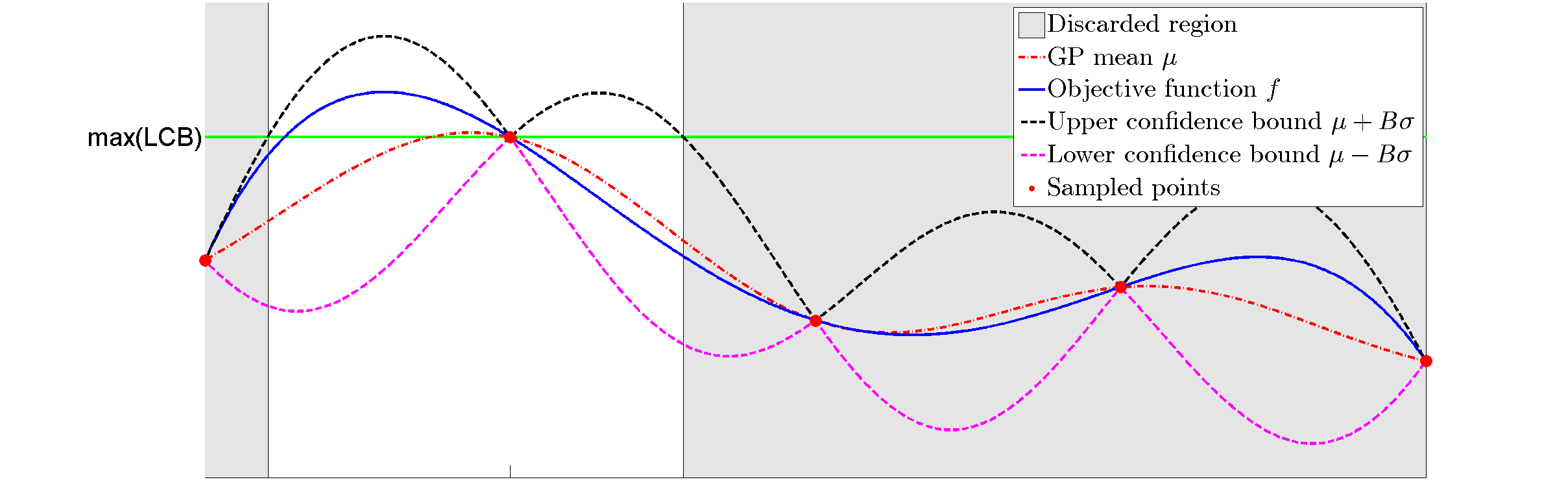}
\end{center}
\caption{\capstyle{An example of our branch and bound maximization algorithm with UCB surrogate $\mu+B \sigma$, where $\mu$ and $\sigma$ are the mean and standard deviation of the GP respectively. The region consisting of the points $x$ for which the upper confidence bound $\mu(x)+B \sigma(x)$ is lower that the maximum value of the lower confidence bound $\mu(x)- B \sigma(x)$ 
 does not need to be sampled anymore. Note that the UCB surrogate function bounds $f$ from above.}} 
\label{fig:BB}
\end{figure*}

This crude approach is wasteful because very often the slope of the function is much smaller than $L$. As we will see below (cf. Figure \ref{fig:BB}), GPs do a better job of providing lower and upper bounds that can be used to limit the search space, by essentially choosing Lipschitz constants that vary over the search space and the algorithm run time.

We also assume that the objective function $f$ is costly to evaluate (e.g. time-wise or financially). We would like to avoid probing $f$ as much as possible and to get close to the optimum as quickly as possible. A solution to this problem is to approximate $f$ with a \emph{surrogate function} that provides a good upper bound for $f$ and which is easier to calculate and optimize. Surrogate functions can also aid with global optimization by restricting the domain of interest.

GPs enable us to construct surrogate functions, which are relatively easy to evaluate and optimize.
We refer the reader to \cite{Brochu2009at} for a general review of the literature on the various surrogate functions utilized in GP bandits in the context of Bayesian optimization.


The surrogate function that we will make extensive use of here is called the Upper Confidence Bound (UCB). It is defined to be $\mu + B\sigma$, where $\mu$ and $\sigma$ are the posterior predictive mean and standard deviation of the GP and $B$ is a constant to be chosen by the algorithm. 
This surrogate function has been studied extensively in the literature and this paper relies heavily on the ideas put forth in the paper by Srinivas et al \cite{Srinivas2010gp}, in which the algorithm consists of repeated optimization of the UCB surrogate function after each sample.

One key difference between our setting and that of \cite{Srinivas2010gp} is that, whereas we assume that the value of the function can be observed exactly, in \cite{Srinivas2010gp} it is necessary for the noise to be non-trivial (and Gaussian) because the main quantity that is used in the estimates, namely information gain, cf. \citep[Equation 3]{Srinivas2010gp}, becomes undefined when the variance of the observation noise ($\sigma^2$ in their notation) is set to $0$, cf. the expression for $\I(\mathbf{y}_A; \mathbf{f}_A)$ that was given in the paragraph following Equation (3). So, their setting is complementary to ours.
Moreover, 
we show that the regret, $r(x_t) = \max_{\mc D} f - f(x_t)$, decreases according to $\mc O\left(e^{-\frac{\tau t}{\left(\ln t\right)^{d/4}}}\right)$, implying that the cumulative regret is bounded from above.


The paper whose results are most similar to ours is \cite{Munos2011soo}, but there are some key differences in the methodology, analysis and obtained rates. For instance, we are interested in cumulative regret, whereas the results of \cite{Munos2011soo} are proven for finite stop-time regret. In our case, the ideal application is the optimization of a function that is $C^2$-smooth and has an unknown non-singular Hessian at the maximum. We obtain a regret rate $\mc O\left(e^{-\frac{\tau t}{\left(\ln t\right)^{d/4}}}\right)$, whereas the DOO algorithm in \cite{Munos2011soo} has regret rate $\mc O(e^{-t})$ if the Hessian is known and the SOO algorithm has regret rate $\mc O(e^{-\sqrt{t}})$ if the Hessian is unknown. In addition, the algorithms in \cite{Munos2011soo} can handle functions that behave like $-c\|x-x_M\|^\alpha$ near the maximum (cf. Example 2 therein). This problem was also studied by
 \cite{Vazquez2011ei} and \cite{Bull2011cr}, but using the 
Expected Improvement surrogate instead of UCB. Our methodology and results are different, but complementary to theirs.

\section{Gaussian process bandits} 
\label{sec:BG}



\subsection{Gaussian processes}
\label{sec:GP}
 
As in \cite{Srinivas2010gp}, the objective function is distributed according to a Gaussian process prior:
\begin{equation}
f(x) \sim \operatorname{GP}(m(\cdot), \kappa(\cdot,\cdot)).
\end{equation}
For convenience, and without loss of generality, we assume that the prior mean vanishes, i.e., $m(\cdot) = 0$.
There are many possible choices for the covariance kernel. One obvious choice is the anisotropic kernel $\kappa$ with a vector of known hyperparameters \cite{Rasmussen2006gp}:
\bea
\kappa(x_i, x_j) &=& \widetilde{\kappa}\left(-(x_i-x_j)^\top\vD(x_i-x_j)\right),
\eea
where $\widetilde{\kappa}$ is an isotropic kernel and $\vD$ is a diagonal matrix with positive hyperparameters along the diagonal and zeros elsewhere. Our results apply to squared exponential kernels and Mat\'ern kernels with parameter $\nu \geq 2$. In this paper, we assume that the hyperparameters are fixed and known in advance.

We can sample the GP at $t$ points by choosing points $\mathbf{x}_{1:t} := \{x_1, \ldots, x_t\}$ and sampling the values of the function at these points to produce the vector $\vf_{1:t} = [f(x_1) \cdots f(x_t)]^\top$. The function values are distributed according to a multivariate Gaussian distribution $\mathcal{N}(0,\vK)$, with covariance entries $ \kappa(x_i, x_j)$.
Assume that we already have several observations from previous steps, and that we want to decide what action $x_{t+1}$ should be considered next. Let us denote the value of the function at this arbitrary new point as $f_{t+1}$. Then, by the properties of GPs, $\vf_{1:t}$ and $f_{t+1}$ are jointly Gaussian:
\[
\begin{bmatrix}
    \vf_{1:t} \\
    f_{t+1}
\end{bmatrix}
\sim {\cal N} 
\left( \mathbf{0} ,
    \begin{bmatrix}
        \vK & \vk^\top \\
        \vk & \kappa(x_{t+1},x_{t+1})
    \end{bmatrix}
\right),
\]
where 
$\vk = [\kappa(x_{t+1},x_1) \cdots \kappa(x_{t+1},x_t)]^\top$. 
Using the Schur complement, one arrives at an expression for the posterior predictive distribution:
\[
P(f_{t+1}|\mathbf{x}_{1:t+1}, \vf_{1:t}) = {\cal N} (\mu_t(x_{t+1}), \sigma_t^2(x_{t+1})),
\]
where
\begin{equation}
\begin{array}{l}
\mu_t(x_{t+1})      = \mathbf{k}^\top \mathbf{K}^{-1} \vf_{1:t},   \\
\sigma_t^2(x_{t+1}) = \kappa(x_{t+1},x_{t+1}) - \mathbf{k}^\top \mathbf{K}^{-1}\mathbf{k}
\end{array}
\label{eqn:Posterior}
\end{equation}
and $\vf_{1:t} = [f(x_1) \cdots f(x_t)]^\top$.

\begin{algorithm*}[t]
\caption{Branch and Bound}
\label{alg:BB}
\begin{algorithmic}
\STATE Input: A compact subset $\mc D \subseteq \bb R^d$, a discrete lattice $\mc L \subseteq \mc D$ and a function $ f: \mc D \to \bb R$.
\STATE $\mc R \gets \mc D$
\STATE $\delta \gets 1$
\REPEAT 
   \STATE \mbox{S\bf ample Twice as Densely:}
   \STATE \qquad $\bullet$ $\delta \gets \dfrac{\delta}{2}$
   \STATE \qquad $\bullet$ Sample $f$ at enough points in $\mc L$ so that every point in $\mc R$ is contained in a simplex of size $\delta$. 
   \STATE \mbox{\bf Shrink the Relevant Region:}
   \STATE \qquad $\bullet$ Set 
   \[ \widetilde{\mc R} := \left\{ x \in \mc R \bigg| \mu_T(x) + \sqrt{\beta_T} \sigma_T(x) > \sup_{\mc R} \mu_T(x) - \sqrt{\beta_T} \sigma_T(x) \right\}. \] 
   \qquad\quad  $T$ is the number points sampled so far and $\beta_T = 2\ln\left(\frac{|\mc L|T^2}{\alpha}\right) = 4\ln T + 2\ln \frac{|\mc L|}{\alpha}$ with $\alpha \in (0,1)$.
   \STATE \qquad $\bullet$ Solve the following constrained optimization problem:
   \[ (x_1^*,x_2^*) = \argsup_{(x_1,x_2) \in \widetilde{\mc R} \times \widetilde{\mc R}} \|x_1 - x_2\| \] 
   \STATE \qquad $\bullet$ $\mc R \gets B\left(\dfrac{x_1^*+x_2^*}{2}, \|x_1^*-x_2^*\|\right)$, where $B(p,r)$ is the ball of radius $r$ centred around $p$.
\UNTIL{$\mc R \cap \mc L = \varnothing$}
\end{algorithmic}
\end{algorithm*}

\subsection{Surrogates for optimization}
When it is assumed that the objective function $f$ is sampled from a GP, one can use a combination of the posterior predictive mean and variance given by Equations~(\ref{eqn:Posterior}) to construct surrogate functions, which tell us where to sample next.
 Here we use the UCB combination, which is given by
\[ 
\mu_t(x) + B_t\sigma_t(x), 
\]
where $\{B_t\}_{t=1}^\infty$ is a sequence of numbers specified by the algorithm. This surrogate trades-off exploration and exploitation since it is optimized by choosing points where the mean is high (exploitation) and where the variance is large (exploration). Since the surrogate has an analytical expression that is easy to evaluate, it is much easier to optimize than the original objective function.
Other popular surrogate functions constructed using the sufficient statistics of the GP include the Probability of Improvement, Expected Improvement and Thompson sampling.
We refer the reader to \cite{Brochu2009at,May2010ob,Hoffman-11} for details on these.

\subsection{Our algorithm}

The main idea of our algorithm (Algorithm \ref{alg:BB}) is to tighten the bound on $f$ given by the UCB surrogate function by sampling the search space more and more densely and shrinking this space as more and more of the UCB surrogate function is ``submerged'' under the maximum of the Lower Confidence Bound (LCB). Figure \ref{fig:BB} illustrates this intuition.

\begin{figure}[t!]
\begin{center}
  \includegraphics[width=0.8\textwidth]{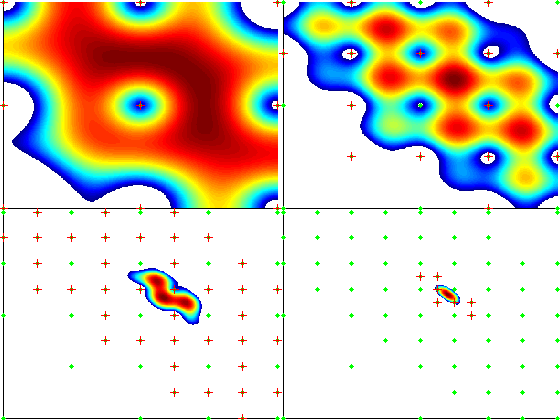}
\end{center}
\caption{\capstyle{Branch and Bound algorithm for a 2D function. The colored region is the search space and the color-map, with red high and blue low, illustrates the value of the UCB. Four steps of the algorithm are shown; progressing from left to right and top to bottom. The green dots designate the points where the function was sampled in the previous steps, while the red crosses denote the freshly sampled points.}} 
\label{fig:BB2D}
\end{figure}

More specifically, the algorithm consists of two iterative stages. During the first stage, the function is sampled along a lattice of points (the red crosses in Figure \ref{fig:BB2D}). In the second stage, the search space is shrunk to discard regions where the maximum is very unlikely to reside. Such regions are obtained by finding points where the UCB is lower than the LCB (the complement of the colored region in the same panel as before). The remaining set of relevant points is denoted by $\widetilde{\mc R}$. In order to simplify the task of shrinking the search space, we simply find an enclosing ball, which is denoted by $\mc R$ in Algorithm \ref{alg:BB}.
Back to the first stage, we consider a lattice that is twice as dense as in the first stage of the previous iteration, but we only sample at points that lie within our new smaller search space.

In the second stage, the auxiliary step of approximating the relevant set $\widetilde{\mc R}$ with the ball $\mc R$ introduces inefficiencies in the algorithm, since we only need to sample inside $\widetilde{\mc R}$. 
This can be easily remedied in practice to obtain an efficient algorithm. Our analysis will show that even without these improvements it is already possible to obtain very strong exponential convergence rates. Of course, practical improvement will result in better constants and ought to be considered seriously.

\section{Analysis}


\subsection{Approximation results}

We begin our analysis by showing that, given sufficient explored
locations, the residual variance is small. More specifically, for any
point $x$ contained in the convex hull of a set of $d$ points that are
no further than $\delta$ apart from $x$, we show that the residual
is bounded by $O(\nbr{h}_\Hcal \delta^2)$, where $\nbr{h}_\Hcal$ is the
Hilbert Space norm of the associated function and that furthermore the
residual variance is bounded by $O(\delta^2)$. 
We begin by relating residual variance, projection operators, and
interpolation in Hilbert Spaces. Lemmas 1, 2 and 3 are standard. We include their proofs in the supplementary material
 for the purpose of being self-contained. Proposition 4 is our key approximation result. It plays a central role in the proof of our exponential regret bounds. Its proof, as well as the proof for the main theorem, is included in the supplementary material.

\begin{lemma}[Hilbert Space Properties]
  \label{lem:rkhs}
  Given a set of points $x_{1:T} := \cbr{x_1, \ldots, x_T} \in \mathcal D$ and a
  Reproducing Kernel Hilbert Space (RKHS) $\Hcal$ with kernel $\kappa$ the following
  bounds hold:
  \begin{enumerate}
  \item \label{item:lipschitz} 
    Any $h \in \Hcal$ is Lipschitz continuous with constant
    $\nbr{h}_\Hcal L$, where $\nbr{\cdot}_\Hcal$ is the Hilbert space norm and $L$ satisfies the following:
    \begin{align}
      L^2 \leq \sup_{x \in \mathcal D} \partial_x \partial_{x'}
      \kappa(x,x')|_{x=x'}
    \end{align}
    and for $\kappa(x,x') = \widetilde{\kappa}(x-x')$ we have
    \[  L^2 \leq \partial_x^2 \widetilde{\kappa}(x)|_{x=0}. \]
  \item \label{item:secondderivative} 
    Any $h \in \Hcal$ has its second derivative bounded by
    $\nbr{h}_\Hcal Q$ where
    \begin{align}
      Q^2 \leq \sup_{x \in \mathcal D} \partial^2_x \partial^2_{x'}
      \kappa(x,x')|_{x=x'}
    \end{align}
      and for $\kappa(x,x') = \widetilde{\kappa}(x-x')$ we have 
      \[ Q^2 \leq \partial_x^4 \widetilde{\kappa}(x)|_{x=0}. \]
  \item \label{item:projection}
    The projection operator $P_{1:T}$ on the subspace $\displaystyle\Span_{t = 1:T} \{\kappa(x_t,\cdot) \} \subseteq \Hcal$ is given by
    \begin{align}
      P_{1:T}h := \vk^\top (\cdot) \vK^{-1} \left< \vk(\cdot), h \right>
    \end{align}
     where $\vk(\cdot) = \vk_{1:T}(\cdot) := \left[\kappa(x_1,\cdot) \cdots \kappa(x_T,\cdot) \right]^\top$ and $\vK := \left[ \kappa(x_i, x_j) \right]_{i,j = 1:T}$; moreover, we have that
     \[ \left< \vk(\cdot), h \right> := \begin{bmatrix} \left< \kappa(x_1, \cdot), h \right> \\ \vdots \\ \left< \kappa(x_T, \cdot), h \right> \end{bmatrix} = \begin{bmatrix} h(x_1) \\ \vdots \\ h(x_T) \end{bmatrix}. \]
    Here $P_{1:T} P_{1:T} = P_{1:T}$ and $\nbr{P_{1:T}} \leq 1$ and $\nbr{\one - P_{1:T}}
    \leq 1$.
  \item \label{item:schurcomplement}
    Given sets $x_{1:T} \subseteq x_{1:T'}$ it follows that $\nbr{P_{1:T} h}_\Hcal \leq
    \nbr{P_{1:T'} h}_\Hcal \leq \nbr{h}_\Hcal$.
  \item \label{item:interpolation}
    Given tuples $(x_i, h_i)$ with $h_i = h(x_i)$, the minimum norm
    interpolation $\bar{h}$ with $\bar{h}(x_i) = h(x_i)$ is given by
    $\bar{h} = P_{1:T} h$. Consequently its residual $g := (\one - P_{1:T}) h$
    satisfies $g(x_i) = 0$ for all $x_i \in x_{1:T}$.
  \end{enumerate}
\end{lemma}

\begin{lemma}[GP Variance]\label{lem:gpvar} Under the assumptions of Lemma~\ref{lem:rkhs} it follows that 
  \begin{align}
    \abr{h(x) - P_{1:T} h(x)} \leq \nbr{h}_\Hcal \sigma_T(x),
  \end{align}
  where $\sigma_T^2(x) = \kappa(x,x) - \vk_{1:T}^\top(x) \vK^{-1} \vk_{1:T}(x)$ 
  and this bound is tight. Moreover, $\sigma_T^2(x)$ is the residual variance
  of a Gaussian process with the same kernel. 
\end{lemma}

\begin{lemma}[Approximation Guarantees]\label{lem:approximation}~We denote by $x_{1:T} \subseteq \mathcal D$ a set of
  locations and assume that $g(x_i) = 0$ for all $x_i \in x_{1:T}$. 
  \begin{enumerate}
  \item Assume that $g$ is Lipschitz continuous with bound $L$. Then
    $g(x) \leq L d(x, x_{1:T})$, where $d(x,x_{1:T})$ is the minimum distance
    $\nbr{x-x_i}$ between $x$ and any $x_i \in x_{1:T}$.
  \item Assume that $g$ has its second derivative bounded by
    $Q^\prime$. Moreover, assume that $x$ is contained inside the convex hull
    of $x_{1:T}$ such that the smallest such convex hull has a maximum
    pairwise distance between vertices of $d$. Then we have 
    $g(x) \leq \frac{1}{4} Q^\prime d^2$.
  \end{enumerate}
\end{lemma}


\begin{proposition}[Variance Bound]
  \label{lem:varbound}
Let $\kappa: \bb R^d \times \bb R^d \to \bb R$ be a kernel that is four times differentiable along the diagonal $\{(x,x) \,|\, x \in \bb R^d\}$, with $Q$ defined as in Lemma \ref{lem:rkhs}.\ref{item:secondderivative}, and $f \sim \operatorname{GP}\left(0,\kappa(\cdot, \cdot)\right)$ a sample from the corresponding Gaussian Process. If $f$ is sampled at points $x_{1:T} = \{x_1, \ldots, x_T\}$ that form a $\delta$-cover of a subset $\mc D \subseteq \bb R^d$, then the resulting posterior predictive standard deviation $\sigma_T$ satisfies
\[ \sup_{\mc D} \sigma_T \leq \frac{Q\delta^2}{4}. \]

\end{proposition}

\subsection{Finiteness of regret}

Having shown that the variance vanishes according to the square of the resolution of the lattice of sampled points, we now move on to show that this estimate implies an exponential asymptotic vanishing of the regret encountered by our Branch and Bound algorithm. This is laid out in our main theorem stated below and proven in the supplementary material.

The theorem considers a function $f$, which is a sample from a GP with a kernel that is four times differentiable along its diagonal. The global maximum of $f$ can appear in the interior of the search space, with the function being twice differentiable at the maximum and with non-vanishing curvature. Alternatively,
the maximum can appear on the boundary with the function having non-vanishing gradient at the maximum.
Given a lattice that is fine enough, the theorem 
asserts that the regret asymptotically decreases in exponential fashion.

The main idea of the proof of this theorem is to use the bound on $\sigma$ given by Proposition~\ref{lem:varbound} to reduce the size of the search space. The key assumption about the function that the proof utilizes is the quadratic upper bound on the objective function $f$ near its global maximum, which together with Proposition~\ref{lem:varbound} allows us to shrink the relevant region $\mc R$ in Algorithm \ref{alg:BB} rapidly. The figures in the proof give a picture of this idea. 
The only complicating factor is the factor $\sqrt{\beta_t}$ in the expression for the UCB that needs to be estimated. This is dealt with by modeling the growth in the number of points sampled in each iteration with a difference equation and finding an approximate solution of that equation.

Recall that $\mc D \subseteq \bb R^d$ is assumed to be a non-empty compact subset and $f$ a sample from the Gaussian Process $\operatorname{GP}\left(0,\kappa(\cdot, \cdot)\right)$ on $\mc D$. Moreover, in what follows we will use the notation $x_M := \displaystyle\argmax_{x \in \mc D} f(x)$. Also, by convention, for any set $\mc S$, we will denote its interior by $\mc S^\circ$, its boundary by $\partial \mc S$ and if $S$ is a subset of $\bb R^d$, then $\cvx(S)$ will denote its convex hull.
The following holds true:
\begin{theorem}\label{thm:BB}
Suppose we are given:
\begin{enumerate}
\item $\alpha > 0$, a compact subset $\mc D \subseteq \bb R^d$,
and $\kappa$ a stationary kernel on $\bb R^d$ that is four times differentiable;
\item $f \sim \operatorname{GP}(0,\kappa)$ a continuous sample on $\mc D$ that has a unique global maximum $x_M$, which satisfies one of the following two conditions:
\begin{itemize}
\item[$(\dagger)$] $x_M \in \mc D^{\circ}$ and $f(x_M) - c_1 \|x-x_M\|^2 < f(x) \leq f(x_M) - c_2 \|x-x_M\|^2$ for all $x$ satisfying $x \in B(x_M, \rho_0)$ for some $\rho_0 > 0$;
\item[$(\ddagger)$] $x_M \in \partial \mc D$ and both $f$ and $\partial \mc D$ are smooth at $x_M$, with $\nabla f(x_M) \neq 0$;
\end{itemize}
\item any lattice $\mc L \subseteq \mc D$ satisfying the following two conditions
\begin{align}
\bullet\quad & 2\mc L \cap \cvx(\mc L) \subseteq \mc L \label{cdn:divby2} \\
\bullet\quad & 2^{\left\lceil -\log_2 \frac{\rho_0}{\diam(\mc D)} \right\rceil + 1}\mc L \cap \mc L \neq \varnothing  \label{cdn:highresolution} \\
        & \text{ if $f$ satisfies $(\dagger)$} \nonumber
\end{align}
\end{enumerate}
Then, there exist positive numbers $A$ and $\tau$ and an integer $T$ such that the points specified by the Branch and Bound algorithm, $\{x_t\}$, will satisfy the following asymptotic bound:
For all $t > T$, with probability $1-\alpha$ we have
\[ r(x_t) < Ae^{-\frac{\tau t}{\left(\ln t\right)^{d/4}}}. \]
\end{theorem}

We would like to make a few clarifying remarks about the theorem. First,
note that for a random sample $f \sim \operatorname{GP}(0,\kappa)$ one of conditions $(\dagger)$ and $(\ddagger)$ will be satisfied almost surely if $\kappa$ is a Mat\'ern kernel with $\nu > 2$ and the squared exponential kernel because the sample $f$ is twice differentiable almost surely by \citep[Theorem 1.4.2]{Adler2007rf} and \citep[\S2.6]{Stein1999gp}) and the vanishing of at least one of the eigenvalues of the Hessian is a co-dimension 1 condition in the space of all functions that are smooth at a given point, so it has zero chance of happening at the global maximum.
Second, the two conditions (\ref{cdn:divby2}) and (\ref{cdn:highresolution}) simply require that the lattice be ``divisible by 2'' and that it be fine enough so that the algorithm can sample inside the ball $B(x_M,\rho_0)$ when the maximum of the function is located in the interior of the search space $\mc D$. Finally, it is important to point out that the rate decay $\tau$ does not depend on the choice of the lattice $\mc L$, even though as stated, the statement of the theorem chooses $\tau$ only after $\mc L$ is specified. The theorem was written this way simply for the sake of readability.

Given the exponential rate of convergence we obtain in Theorem \ref{thm:BB}, we have the following finiteness conclusion for the cumulative regret accrued by our Branch and Bound algorithm:
\begin{corollary} Given $\kappa$, $f \sim \operatorname{GP}(0,\kappa)$ and $\mc L \subseteq \mc D$ as in Theorem \ref{thm:BB}, the cumulative regret is bounded from above. 
\end{corollary}

\begin{remark} It is worth pointing out the trivial observation that using a simple UCB algorithm with monotonically increasing and unbounded factor $\sqrt{\beta_t}$, without any shrinking of the search space as we do here, necessarily leads to unbounded cumulative regret since eventually $\sqrt{\beta_t}$ becomes large enough so that at points $x^\prime$ far away from the maximum, $\sqrt{\beta_t}\sigma_t(x^\prime)$ becomes larger than $f(x_M)-f(x)$. In fact, eventually the UCB algorithm will sample every point in the lattice $\mc L$.
\end{remark}

\section{Discussion}

In this paper we proposed a modification of the UCB algorithm of
\cite{Srinivas2010gp} which addresses the noise free case. The key
difference is that while the original algorithm achieves an
$O(t^{-\frac{1}{2}})$ rate of convergence to the regret minimizer, we
obtain an exponential rate in the number of function evaluations. In
other words, the noise free problem is significantly easier,
statistically speaking, than the noisy case. The key difference is
that we need not invest any samples in noise reduction to determine
whether our observations deviate far from their expectation.

This allows us to discard pieces of the search space where the maximum
is very unlikely to be, when compared to \cite{Srinivas2010gp}. We
show that this additional step leads to a considerable improvement of
the regret accrued by the algorithm. In particular, the cumulative
regret obtained by our Branch and Bound algorithm is bounded from
above, whereas the cumulative regret bound obtained in the noisy
bandit algorithm is unbounded. The possibility of dispensing with
chunks of the search space can also be seen in the works involving
hierarchical partitioning, e.g. \cite{Munos2011soo}, where regions of
the space are deemed as less worthy of probing as time goes on.

Our results mirror the observation in active learning that noise free
and large margin learning of half spaces can be achieved much more
rapidly than identifying a linear separator in the noisy case
\cite{BshWat06,DasKalMon09}. This is also reflected in classical
uniform convergence results for supervised learning
\cite{AudTsy07,Vapnik98} where the achievable rate depends on the
decay of probability mass near the margin.

This suggests that the ability to extend our results to the noisy case
is somewhat limited. An indication of what might be possible can be
found in \cite{BalBeyLan09}, where regions of the version space are
eliminated once they can be excluded with sufficiently high
probability. One could model a corresponding Branch and Bound
algorithm, which dispenses with points that lie outside the current
(or perhaps the previous) relevant set when calculating the covariance
matrix $\mathbf K$ in the posterior equations
(\ref{eqn:Posterior}). Analysis of how much of an effect such a
computational cost-cutting measure would have on the regret
encountered by the algorithm is a subject of future research. 


We believe that an exciting extension can be found in guarantees for
contextual bandits. Note, however, that the unpredictability of the
context introduces new difficulties in terms of speed of convergence
that need to be overcome. For instance, parameters for infrequent 
contexts will be estimated slowly unless there are strong correlations among contexts.

\bibliography{BayesBandits}

\begin{thebibliography}{27}
\providecommand{\natexlab}[1]{#1}
\providecommand{\url}[1]{\texttt{#1}}
\expandafter\ifx\csname urlstyle\endcsname\relax
  \providecommand{\doi}[1]{doi: #1}\else
  \providecommand{\doi}{doi: \begingroup \urlstyle{rm}\Url}\fi

\bibitem[Adler \& Taylor(2007)Adler and Taylor]{Adler2007rf}
Adler, Robert~J. and Taylor, Jonathan~E.
\newblock \emph{Random Fields and Geometry}.
\newblock Springer, 2007.

\bibitem[Audibert \& Tsybakov(2007)Audibert and Tsybakov]{AudTsy07}
Audibert, Jean-Yves and Tsybakov, Alexandre~B.
\newblock Fast learning rates for plug-in classifiers.
\newblock \emph{Annals of Statistics}, 35\penalty0 (2):\penalty0 608--633,
  2007.

\bibitem[Balcan et~al.(2009)Balcan, Beygelzimer, and Langford]{BalBeyLan09}
Balcan, Maria-Florina, Beygelzimer, Alina, and Langford, John.
\newblock Agnostic active learning.
\newblock \emph{J. Comput. Syst. Sci}, 75\penalty0 (1):\penalty0 78--89, 2009.

\bibitem[Brochu et~al.(2007)Brochu, Freitas, and Ghosh]{Brochu07}
Brochu, Eric, Freitas, Nando~De, and Ghosh, Abhijeet.
\newblock Active preference learning with discrete choice data.
\newblock In \emph{Advances in Neural Information Processing Systems}, pp.\
  409--416, 2007.

\bibitem[Brochu et~al.(2009)Brochu, Cora, and {de Freitas}]{Brochu2009at}
Brochu, Eric, Cora, Vlad~M, and {de Freitas}, Nando.
\newblock A tutorial on {B}ayesian optimization of expensive cost functions,
  with application to active user modeling and hierarchical reinforcement
  learning.
\newblock Technical Report TR-2009-023, arXiv:1012.2599v1, UBC CS department,
  2009.

\bibitem[Bshouty \& Wattad(2006)Bshouty and Wattad]{BshWat06}
Bshouty, Nader~H. and Wattad, Ehab.
\newblock On exact learning halfspaces with random consistent hypothesis
  oracle.
\newblock In \emph{International Conference on Algorithmic Learning Theory},
  pp.\  48--62, 2006.

\bibitem[Bubeck et~al.(2011)Bubeck, Munos, Stoltz, and Szepesvari]{Xarmed2011}
Bubeck, S\'ebastien, Munos, R\'emi, Stoltz, Gilles, and Szepesvari, Csaba.
\newblock X-armed bandits.
\newblock \emph{Journal of Machine Learning Research}, 12:\penalty0 1655--1695,
  2011.

\bibitem[Bull(2011)]{Bull2011cr}
Bull, Adam~D.
\newblock Convergence rates of efficient global optimization algorithms.
\newblock \emph{Journal of Machine Learning Research}, 12:\penalty0 2879--2904,
  2011.

\bibitem[Dasgupta et~al.(2009)Dasgupta, Kalai, and Monteleoni]{DasKalMon09}
Dasgupta, Sanjoy, Kalai, Adam~Tauman, and Monteleoni, Claire.
\newblock Analysis of perceptron-based active learning.
\newblock \emph{Journal of Machine Learning Research}, 10:\penalty0 281--299,
  2009.

\bibitem[Garnett et~al.(2010)Garnett, Osborne, and Roberts]{Garnett-10}
Garnett, R., Osborne, MA, and Roberts, SJ.
\newblock {Bayesian optimization for sensor set selection}.
\newblock In \emph{ACM/IEEE International Conference on Information Processing
  in Sensor Networks}, pp.\  209--219. ACM, 2010.

\bibitem[Ghosal \& Roy(2006)Ghosal and Roy]{GhosalRoy2006}
Ghosal, Subhashis and Roy, Anindya.
\newblock Posterior consistency of {Gaussian} process prior for nonparametric
  binary regression.
\newblock \emph{Ann. Stat.}, 34:\penalty0 2413--2429, 2006.

\bibitem[Gramacy et~al.(2004)Gramacy, Lee, and MacReady]{Gramacy-04}
Gramacy, Robert~B., Lee, Herbert K.~H., and MacReady, William.
\newblock Parameter space exploration with {Gaussian} process trees.
\newblock In \emph{International Conference on Machine Learning}, pp.\
  353--360, 2004.

\bibitem[Hansen et~al.(1992)Hansen, Jaumard, and Lu]{LipschitzReview1992}
Hansen, P., Jaumard, B., and Lu, S.
\newblock Global optimization of univariate {Lipschitz} functions: I. survey
  and properties.
\newblock \emph{Mathematical Programming}, 55:\penalty0 251--272, 1992.

\bibitem[Hoffman et~al.(2011)Hoffman, Brochu, and de~Freitas]{Hoffman-11}
Hoffman, Matthew, Brochu, Eric, and de~Freitas, Nando.
\newblock Portfolio allocation for {Bayesian} optimization.
\newblock In \emph{Uncertainty in Artificial Intelligence}, pp.\  327--336,
  2011.

\bibitem[Hutter et~al.(2010)Hutter, Hoos, and Leyton-Brown]{Hutter2010mip}
Hutter, Frank, Hoos, Holger~H., and Leyton-Brown, Kevin.
\newblock Automated configuration of mixed integer programming solvers.
\newblock In \emph{Proceedings of CPAIOR-10}, pp.\  186–--202, 2010.

\bibitem[Lizotte(2008)]{Lizotte-08}
Lizotte, Daniel.
\newblock \emph{Practical {Bayesian} Optimization}.
\newblock PhD thesis, University of Alberta, Edmonton, Alberta, Canada, 2008.

\bibitem[{Martinez--Cantin} et~al.(2009){Martinez--Cantin}, {de Freitas},
  Brochu, Castellanos, and Doucet]{Martinez-Cantin-09}
{Martinez--Cantin}, Ruben, {de Freitas}, Nando, Brochu, Eric, Castellanos,
  Jose, and Doucet, Arnaud.
\newblock A {Bayesian} exploration-exploitation approach for optimal online
  sensing and planning with a visually guided mobile robot.
\newblock \emph{Autonomous Robots}, 27\penalty0 (2):\penalty0 93--103, 2009.

\bibitem[May et~al.(2010)May, Korda, Lee, and Leslie]{May2010ob}
May, Benedict, Korda, Nathan, Lee, Anthony, and Leslie, David.
\newblock Optimistic {Bayesian} sampling in contextual-bandit problems.
\newblock 2010.

\bibitem[Mo{\v c}kus(1982)]{Mockus-82}
Mo{\v c}kus, Jonas.
\newblock The {B}ayesian approach to global optimization.
\newblock In \emph{System Modeling and Optimization}, volume~38, pp.\
  473--481. Springer Berlin / Heidelberg, 1982.

\bibitem[Munos(2011)]{Munos2011soo}
Munos, R\'emi.
\newblock Optimistic optimization of a deterministic function without the
  knowledge of its smoothness.
\newblock In \emph{Advances in Neural Information Processing Systems}, 2011.

\bibitem[Rasmussen \& Williams(2006)Rasmussen and Williams]{Rasmussen2006gp}
Rasmussen, Carl~Edward and Williams, Christopher K.~I.
\newblock \emph{Gaussian Processes for Machine Learning}.
\newblock The MIT Press, 2006.

\bibitem[Schonlau et~al.(1998)Schonlau, Welch, and Jones]{Schonlau-98}
Schonlau, Matthias, Welch, William~J., and Jones, Donald~R.
\newblock Global versus local search in constrained optimization of computer
  models.
\newblock \emph{Lecture Notes-Monograph Series}, 34:\penalty0 11--25, 1998.

\bibitem[Srinivas et~al.(2010)Srinivas, Krause, Kakade, and
  Seeger]{Srinivas2010gp}
Srinivas, Niranjan, Krause, Andreas, Kakade, Sham~M, and Seeger, Matthias.
\newblock Gaussian process optimization in the bandit setting: No regret and
  experimental design.
\newblock In \emph{International Conference on Machine Learning}, 2010.

\bibitem[Stein(1999)]{Stein1999gp}
Stein, Michael~L.
\newblock \emph{Interpolation of Spatial Data: Some Theory for Kriging}.
\newblock Springer, 1999.

\bibitem[Steinwart \& Christmann(2008)Steinwart and
  Christmann]{Steinwart2008svm}
Steinwart, Ingo and Christmann, Andreas.
\newblock \emph{Support Vector Machines}.
\newblock Springer, 2008.

\bibitem[Vapnik(1998)]{Vapnik98}
Vapnik, V.
\newblock \emph{Statistical Learning Theory}.
\newblock John Wiley and Sons, New York, 1998.

\bibitem[Vazquez \& Bect(2010)Vazquez and Bect]{Vazquez2011ei}
Vazquez, Emmanuel and Bect, Julien.
\newblock Convergence properties of the expected improvement algorithm with
  fixed mean and covariance functions.
\newblock \emph{Journal of Statistical Planning and Inference}, 140:\penalty0
  3088--3095, 2010.

\end{thebibliography}
\bibliographystyle{techreport}

\section{Proofs}
\subsection{Approximation Results}

\begin{proof}[Lemma \ref{lem:rkhs}]
  We prove the claims in sequence. 
  \begin{enumerate}
  \item This follows from Corollary 4.36 in \cite{Steinwart2008svm}, with 
    $|\alpha| = 1$.
  \item Same as above, just with $|\alpha| = 2$.
\oneoftwo{\item For any operator $V$ with full column rank the projection on
    the image of $V$ is given by $V (V^\top V)^{-1} V^\top$. The operator
    $V$ in the above case is given by the stacked vector of evaluation
    functionals $k(x_1, \cdot), \ldots, k(x_n, \cdot)$. This provides
    us with $P_X$. The remaining claims are standard linear algebra.

}{
  \item For any operator $V$ with full column rank the projection on
    the image of $V$ is given by $V (V^\top V)^{-1} V^\top$. In our case, $V$ is defined as
\[ \mapdefDisp{V}{\bb R^T}{\Hcal}{\vw := \mat{ w_1 \\ \vdots \\ w_T }}{\vk^\top(\cdot)\vw}. \]

To calculate the adjoint $V^\top$, let $h \in \Hcal$ and $\vw \in \bb R^T$ be arbitrary elements and consider the following chain of equalities:
\begin{align*}
\left< V^\top h, \vw \right>_{\bb R^T} & = \left< h, V\vw \right>_\Hcal \qquad\qquad \text{(cf. \cite{Steinwart2008svm} Equation A.19)} \\
                             & = \left< h, \vk^\top(\cdot)\vw \right>_\Hcal \\
                             & = \left< h, \mat{k(x_1,\cdot) & \cdots & k(x_T,\cdot)} \mat{w_1 \\ \vdots \\ w_T} \right>_\Hcal \\
                             & = \left< h, k(x_1,\cdot)w_1 + \cdots + k(x_T,\cdot)w_T \right>_\Hcal \\
                             & = \left< h, k(x_1,\cdot) \right> w_1 + \cdots + \left< h, k(x_T,\cdot) \right> w_T \qquad \text{(by linearity of $\left< \, , \right>_\Hcal$)}\\
                             & = \mat{\left< h, k(x_1,\cdot)\right>_\Hcal & \cdots & \left< h, k(x_T,\cdot)\right>_\Hcal} \mat{w_1 \\ \vdots \\ w_T} \\
                             & = \left< \vk(\cdot), h \right>_\Hcal^\top \vw \;\,\qquad\qquad \text{(by the symmetry of $\left< \, , \right>_\Hcal$)}\\
                             & = \left< \left< \vk(\cdot), h \right>_\Hcal, \vw \right>_{\bb R^T} \qquad \text{(by the definition of $\left< \, , \right>_{\bb R^T}$)}
\end{align*}
Since, this equality holds for all $\vw$, we can conclude that for all $h \in \Hcal$
\[ V^\top h = \left< \vk(\cdot), h \right>_\Hcal. \]

Now, all we need to do is to calculate the expression $V^\top V$ to complete the derivation of the expression for $P_{1:T}$; to this end let $\vw \in \bb R^T$ be arbitrary:
\begin{align*}  
V^\top V \vw & = \left< \vk(\cdot), \vk^\top(\cdot) \vw \right>_\Hcal \\
             & = \left< \vk(\cdot), \vk^\top(\cdot) \right>_\Hcal \vw \\
             & = \left< \mat{\kappa(x_1,\cdot) \\ \vdots \\ \kappa(x_T,\cdot)}, \mat{\kappa(x_1,\cdot) & \cdots & \kappa(x_T,\cdot)} \right>_\Hcal \vw \\
             & = \begin{bmatrix} 
\left< \kappa(x_1,\cdot), \kappa(x_1,\cdot) \right>_\Hcal & \cdots & \left< \kappa(x_1,\cdot), \kappa(x_T,\cdot) \right>_\Hcal \\
\vdots & \ddots & \vdots \\
\left< \kappa(x_T,\cdot), \kappa(x_1,\cdot) \right>_\Hcal & \cdots & \left< \kappa(x_T,\cdot), \kappa(x_T,\cdot) \right>_\Hcal                 
\end{bmatrix} \vw \\
             & = \begin{bmatrix} 
\kappa(x_1,x_1) & \cdots & \kappa(x_1,x_T)\\
\vdots & \ddots & \vdots \\
\kappa(x_T,x_1) & \cdots & \kappa(x_T,x_T)                 
\end{bmatrix} \vw \qquad \text{(cf. \cite{Steinwart2008svm} Definition 4.18 and Equation 4.14)} \\
             & = \vK \vw
\end{align*}
and so $V^\top V = \vK$.


This provides us with $P_{1:T}$. The remaining claims follow from standard properties of projection operators.}

  \item Projection operators satisfy $\nbr{P_{1:T}} \leq 1$. This proves
    the second claim. The first claim can be seen from the fact that
    projecting on a subspace can only have a smaller norm than the
    superspace projection. 
  \item We first show that the projection is an interpolation. This
    follows from
    \begin{align*}
      \bar{h}(x_i) = P_{1:T} h(x_i) = \inner{P_{1:T} h}{\kappa(x_i, \cdot)}
      = \inner{h}{P_{1:T} \kappa(x_i, \cdot)}     = \inner{h}{\kappa(x_i, \cdot)} =
      h(x_i).
    \end{align*}
    Correspondingly $g(x_i) = h(x_i) - \bar{h}(x_i) = 0$ for all $x_i
    \in x_{1:T}$. By construction $P_{1:T} h$ uses $h$ only in evaluations
    $h(x_i)$, hence for any two functions $h, h'$ with $h(x_i) =
    h'(x_i)$ we have $P_{1:T} h = P_{1:T} h'$. Since $\nbr{P_{1:T}} \leq
    1$ it follows that $\nbr{P_{1:T} h} \leq \nbr{h}_\Hcal$. Hence there is no
    interpolation with norm smaller than $\nbr{P_{1:T} h}$.
  \end{enumerate}
\end{proof}

\begin{proof}[Lemma \ref{lem:gpvar}]
  To see the bound we again use the Cauchy-Schwartz inequality
  \begin{align*}
    \abr{h(x) - P_{1:T} h(x)} & = 
    \abr{(\one - P_{1:T}) h(x)} \\ 
    & = \abr{\inner{(\one - P_{1:T}) h}{\kappa(x,\cdot)}_\Hcal} \quad \text{(by the defining property of $\left<\, , \right>_\Hcal$,} \\
    & \qqquad\qqquad\qqquad\quad\; \text{cf. \cite{Steinwart2008svm}, Def. 4.18)} \\
    & = \abr{\inner{h}{(\one - P_{1:T}) \kappa(x,\cdot)}_\Hcal} \quad \text{(since $\one - P_{1:T}$ is an orthogonal projection and so self-adjoint)} \\
    & \leq \nbr{h}_\Hcal \nbr{(\one - P_{1:T}) \kappa(x,\cdot)} \quad\; \text{(by Cauchy-Schwarz)}
  \end{align*}

  This inequality is clearly tight for $h = (\one - P_{1:T}) \kappa(x,
  \cdot)$ by the nature of dual norms. Next note that 
  \begin{align*}
    \nbr{(\one - P_{1:T}) \kappa(x, \cdot)}^2 & =
    \inner{(\one - P_{1:T}) \kappa(x, \cdot)}{(\one - P_{1:T}) \kappa(x, \cdot)} =
    \inner{\kappa(x, \cdot)}{(\one - P_{1:T}) \kappa(x, \cdot)} \\
    & = \kappa(x,x) - \inner{\kappa(x, \cdot)}{P_{1:T} \kappa(x, \cdot)} = \sigma_T^2(x).
  \end{align*}
  The second equality follows from the fact that $\one - P_{1:T}$ is
  idempotent. The last equality follows from the definition of
  $P_{1:T}$. The fact that $\sigma_T^2(x)$ is the residual variance of a
  Gaussian Process regression estimate is well known in the literature
  and follows, e.g.\ from the matrix inversion lemma.
\end{proof}

\begin{proof}[Lemma \ref{lem:approximation}]
  The first claim is an immediate consequence of the Lipschitz
  property of $g$. To see the second claim we need to establish a
  number of issues: without loss of generality assume that the maximum
  within the convex hull containing $x$ is attained at $x$ (and that
  the maximum rather than the minimum denotes the maximum deviation
  from $0$). 

  The maximum distance of $x$ to one of its vertices is bounded by
  $\delta/\sqrt{2}$. This is established by considering the minimum
  enclosing ball and realizing that the maximum distance is achieved
  for the regular polyhedron. 

  To see the maximum deviation from $0$ we exploit the fact that
  $\partial_x g(x) = 0$ by the assumption of $x$ being the maximum (we
  need not consider cases where $x$ is on a facet of the polyhedral
  set since in this case we could easily reduce the
  dimensionality). In this case the largest deviation between $g(x)$
  and $g(x_i)$ is obtained by making $g$ a quadratic function $g(x') =
  \frac{Q^\prime}{2} \nbr{x'-x}^2$. At distance $\frac{\delta}{\sqrt{2}}$ the
  function value is bounded by $\frac{\delta^2 Q^\prime}{4}$. Since the latter
  bounds the maximum deviation it does bound it for $g$ in
  particular. This proves the claim. 
\end{proof}

\begin{proof}[Proposition \ref{lem:varbound}]
Let $\Hcal$ be the RKHS corresponding to $\kappa$ and $h \in \Hcal$ an arbitrary element, with $g := (\one - P_{1:T}) h$ the residual defined in Lemma \ref{lem:rkhs}.\ref{item:interpolation}. By Lemma \ref{lem:rkhs}.\ref{item:projection}, we know that $\nbr{\one - P_{1:T}} \leq 1$ and so we have
\begin{equation}
 \nbr{g}_\Hcal \leq \nbr{\one - P_{1:T}}\nbr{h}_\Hcal \leq \nbr{h}_\Hcal \label{eqn:proj}
\end{equation}

Moreover, by Lemma \ref{lem:rkhs}.\ref{item:secondderivative}, we know that the second derivative of $g$ is bounded by $\nbr{g}_\Hcal Q$, and since by Lemma \ref{lem:rkhs}.\ref{item:interpolation} we know that $g$ vanishes at each $x_i$, we can use Lemma \ref{lem:approximation}.2 and the inequality given by inequality (\ref{eqn:proj}) to conclude that
\begin{align*}
|h(x) - P_{1:T}h(x)| & := |g(x)| \\
   & \leq \frac{\nbr{g}_\Hcal Q\delta^2}{4} \; \text{by Lemma \ref{lem:approximation}.2} \\
   & \leq \frac{\nbr{h}_\Hcal Q\delta^2}{4} \; \text{by inequality (\ref{eqn:proj})}
\end{align*}
and so for all $x \in \mc D$ we have
\begin{equation}
|h(x) - P_{1:T}h(x)| \leq \frac{Q\delta^2}{4}\nbr{h}_\Hcal  \label{ineq:CurvBound}
\end{equation}

On the other hand, by Lemma \ref{lem:gpvar}, we know that for all $x \in \mc D$ we have the following tight bound:
\begin{equation}
|h(x) - P_{1:T}h(x)| \leq \sigma_T(x) \nbr{h}_\Hcal. \label{ineq:CSch}
\end{equation}

Now, given the fact that both inequalities (\ref{ineq:CurvBound}) and (\ref{ineq:CSch}) are bounding the same quantity and that the latter is a tight estimate, we necessarily have that
\[ \sigma_T(x) \nbr{h}_\Hcal \leq \frac{Q\delta^2}{4}\nbr{h}_\Hcal. \]
Canceling $\nbr{h}_\Hcal$ gives the desired result.

\end{proof}

\subsection{Finiteness of Regret}
We begin with two lemmas from \cite{Srinivas2010gp}:
\begin{lemma}[Lemma 5.1 of \cite{Srinivas2010gp}]\label{lem:SKKS5.1} Given any finite set $\mc L$, any sequence of points $\{x_1, x_2, \ldots\} \subseteq \mc L$ and $f: \mc L \to \bb R$ a sample from $\operatorname{GP}(0, \kappa(\cdot,\cdot))$, for all $\alpha \in (0,1)$, we have
\[  P\left\{ \forall x \in \mc L, t \geq 1: \; |f(x) - \mu_{t-1}(x)| \leq \sqrt{\beta_t}\sigma_{t-1}(x) \right\} \geq 1-\alpha, \]
where $\beta_t = 2\ln\left(\frac{|\mc L|\pi_t}{\alpha}\right)$ and $\{\pi_t\}$ is any positive sequence satisfying $\displaystyle\sum_t \frac{1}{\pi_t} = 1$. Here $|\mc L|$ denotes the number of elements in $\mc L$.
\end{lemma}

\begin{lemma}[Lemma 5.2 in \cite{Srinivas2010gp}]\label{lem:SKKS5.2} Let $\mc L$ a non-empty finite set and $f: \mc L \to \bb R$ an arbitrary function. Also assume that there exist functions $\mu, \sigma: \mc L \to \bb R$ and a constant $\sqrt{\beta}$, such that 
\begin{equation}
 |f(x) - \mu(x)| \leq \sqrt{\beta}\sigma \quad \forall\, x \in \mc L.  \label{cdn:Lemma5.2} 
\end{equation}
Then, we have
\[ r(x) \leq 2\sqrt{\beta}\sigma(x) \leq 2\sqrt{\beta}\max_{\mc L} \sigma. \]
\end{lemma}

\begin{definition}[Covering Number]
  \label{def:cover}
  Denote by $\Bcal$ a Banach space with norm
  $\nbr{\cdot}$. Furthermore denote by $B \subseteq \Bcal$ a set in
  this space. Then the covering number $n_\epsilon(B, \Bcal)$ is defined as
  the minimum number of $\epsilon$ balls with respect to the Banach
  space norm that are required to cover $B$ entirely. 
\end{definition}

\begin{proof}[Theorem \ref{thm:BB}]
The proof consists of the following steps:
\begin{itemize}
\item[$\bullet$]{\bf Global:} We first show that after a finite number of steps the algorithm zooms in on the neighbourhood $B(x_M,\rho_0)$. This is done by first showing that $\epsilon$ can be chosen small enough to squeeze the set $f^{-1}((f_M-\epsilon,f_M])$ into any arbitrarily small neighbourhood of $x_M$ and that as the function is sampled more and more densely, the UBC-LCB envelope around $f$ becomes arbitrarily tight, hence eventually fitting the relevant set inside a small neighbourhood of $x_M$. Please refer to Figure \ref{fig:ProofGlobal} for a graphical depiction of this process.
\begin{itemize}
\item[G$_I$:] Since $\mc D$ is compact and $f$ is continuous and has a unique maximum, for every $\rho > 0$, we can find an $\epsilon = \epsilon(\rho) > 0$ such that 
\[ f^{-1}\left((f_M-\epsilon,f_M]\right) \subseteq B(x_M, \rho), \]
where $f_M = \max f$.

To see this, suppose on the contrary that there exists a radius $\rho > 0$ such that for all $\epsilon > 0$ we have 
\[ f^{-1}\left((f_M-\epsilon,f_M]\right) \nsubseteq B(x_M, \rho) \]
which means that there exists a point $x \in \mc D$ such that $f(x_M)-f(x) < \epsilon$ but $\|x-x_M\| > \rho$. Now, for each $i \in \bb N$, pick a point $x^i \in f^{-1}\left((f_M-\frac{1}{i},f_M]\right) \setminus B(x_M, \rho)$: this gives us a sequence of points $\{x^i\}$ in $\mc D$, which by the compactness of $\mc D$ has a convergent subsequence $\{x^{i_k}\}$, whose limit we will denote by $x^*$. From the continuity of $f$ and the fact that $f(x_M) - f(x^i) < \frac{1}{i}$, we can conclude that $f(x_M) - f(x^*) = 0$, which contradicts our assumption that $f$ has a unique global maximum since we necessarily have $x^* \notin B(x_M,\rho)$.

\begin{figure*}[tb]
\begin{center}
  \includegraphics[width=\textwidth]{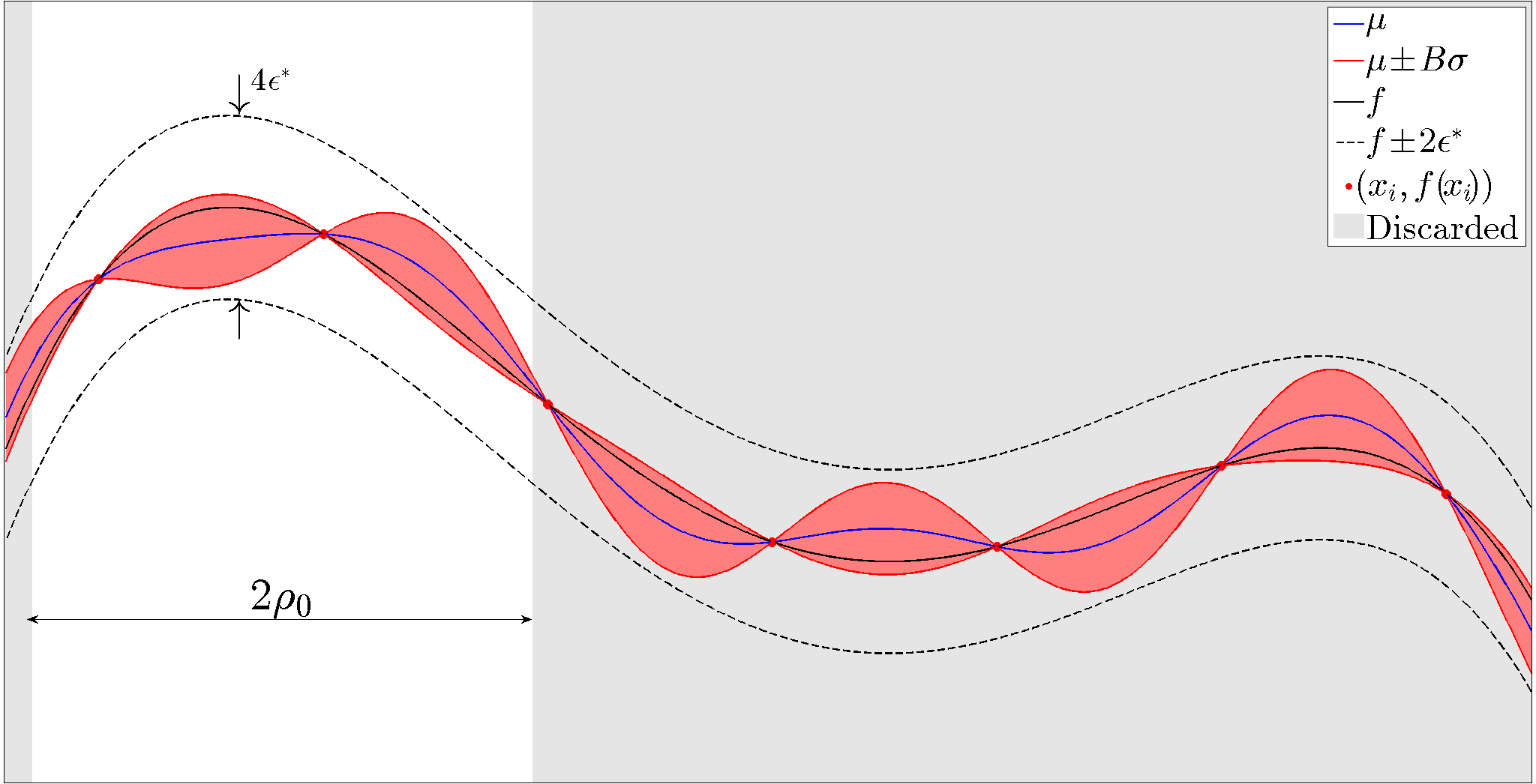}
\end{center}
\caption{\capstyle{The elimination of other smaller peaks.}} 
\label{fig:ProofGlobal}
\end{figure*}

\item[G$_{II}$:] Define $\epsilon^* := \dfrac{\epsilon(\rho_0)}{4}$, with $\rho_0$ as in Condition $(\dagger)$ of the statement of Theorem \ref{thm:BB}.
\item[G$_{III}$:] For each $T$, define the ``relevant set'' $\mc R_T \subseteq \mc D$ as follows:
\[ \mc R_T = \left\{ x \in \mc D \bigg| \mu_T(x) + \sqrt{\beta_T} \sigma_T(x) > \sup_{\mc R} \mu_T(x) - \sqrt{\beta_T} \sigma_T(x) \right\}. \]
\item[G$_{IV}$:] Choose $\beta_T = b\ln(T)$, with $b$ chosen large enough to satisfy the conditions of Lemma \ref{lem:SKKS5.1}. Then, it is possible to sample $f$ densely enough so that 
\begin{equation}
\sqrt{\beta_T} \max_{x \in \mc D} \sigma_T(x) < \epsilon^*,  \label{ineq:sigmaBound}
\end{equation}
so that $\mc R_T \subseteq B(x_M, \rho_0)$. This is because as $\mc D$ is sampled more and more densely we have $\sigma = O(\delta^2)$, where $\delta$ is the distance between the points of the grid, and $\beta = O\left(\ln\frac{1}{\delta^d}\right) = O\left(-\ln \delta\right)$ and so $\sqrt{\beta}\sigma \to 0$ as $\delta \to 0$, and so there exists a $\delta_0$ small enough so that a lattice of resolution $\delta_0$ would give us the bound given in inequality  $(\ref{ineq:sigmaBound})$. The end point of this process is depicted in Figure \ref{fig:ProofGlobal}, where the relevant set $\mc R_T$ lies inside the non-shaded region: the reason for this inclusion and ``thickness'' $4\epsilon^*$ is described below, in Step L$_1$ of the proof: cf. Equation (\ref{eqn:fMuSigmaSandwich}).

\end{itemize}
\item[$\bullet$]{\bf Local:} Once the algorithm has localized attention to a neighbourhood of $x_M$, then we can show that the regret decreases exponentially; to do so, we will proceed by sampling the relevant set twice as densely and shrinking the relevant set, and repeating these two steps. The claim is that in each iteration, the maximum regret goes down exponentially and the number of the new points that are sampled in each {\bf refining} iteration is asymptotically constant. To prove this, we will write down the equations governing the behaviour of the number of sampled points and $\sigma$. We will adopt the following notation to carry out this task:
\begin{itemize}
\item $\delta_\ell$ - the resolution of the lattice of sampled points at the end of the $(\ell+1)^{th}$ refining iteration inside $\mc R_{\ell+1}$ (defined below).
\item $\epsilon_\ell = \displaystyle\sup_{x \in \mc R_\ell} \sigma_{N_\ell}(x)$ at the end of the $\ell^{th}$ iteration. Note that $\epsilon_\ell \propto \delta_\ell^2$. Also, note that $\epsilon_0 \leq \epsilon^*$ by the choice of $\delta_0$.
\item $N_\ell$ - number of points that have been sampled by the end of the $\ell^{th}$ iteration.
\item $\Delta N_\ell = N_{\ell+1} - N_\ell$.
\item $\mc R_\ell$ - the relevant set at the beginning of the $\ell^{th}$ iteration. Note that $\mc R_1 \subseteq B(x_M, \rho_0)$.
\item $\rho_\ell = \dfrac{\diam(\mc R_\ell)}{2}$. Note that $\rho_1 < \rho_0$.
\end{itemize}

\begin{itemize}
\item[L$_1$:] 
\begin{align*}
 N_1 & \leq N_0 + n_{\delta_0}\left(\mc R_0, (\bb R^d, \|\cdot\|_2) \right) \quad \text{where $n_{\delta_0}\left(\mc R_0, (\bb R^d, \|\cdot\|_2) \right)$ is the $\delta_0$-covering number} \\ 
     & \qqquad\qqquad\qquad\qquad\qquad\, \text{as defined in Definition \ref{def:cover}} \\
     & \leq N_0 + \mc N(\rho_0, \delta_0) \quad\qquad\;\, \text{where $\mc N(\rho_0, \delta_0) := n_{\delta_0}\left(B(0, \rho_0), (\bb R^d,\|\cdot\|_2) \right)$} \\
     & \leq N_0 + \mc N\left(\sqrt{\frac{4\epsilon_0 \sqrt{\beta_{N_0}}}{c_2}}, \delta_0\right) \\
     & \leq N_0 + \mc N\left(\sqrt{\frac{4\epsilon_0 \sqrt{b\ln N_0}}{c_2}}, \delta_0\right) \\
     & = N_0 + \mc N\left(c\sqrt{\epsilon_0} \sqrt[4]{\ln N_0}, \delta_0\right) \qquad \text{where $c := \sqrt{\frac{4\sqrt{b}}{c_2}}$}
\end{align*}

The expression $\sqrt{\frac{4\epsilon_0 \sqrt{\beta_{N_0}}}{c_2}}$ comes about as follows: using the notations $B = \sqrt{\beta_{\!N_0}}$ and $\sigma = \sigma_{N_0}$ we know by Lemma \ref{lem:SKKS5.1} that $f$ and $\mu$ are intertwined with each other in the sense that both of the following chains of inequality hold:
\begin{align*}
 \mu\!-\!\!B\sigma \quad\leq\quad & f \quad\leq\quad \mu\!+\!\!B\sigma \\
 f\!-\!\!B\sigma \quad\leq\quad & \mu \quad\leq\quad f\!+\!\!B\sigma,
\end{align*}
which, combined together, give us the following chain of inequalities
\begin{equation}\label{eqn:fMuSigmaSandwich}
 f\!-\!\!2B\sigma \quad\leq\quad \mu\!-\!\!B\sigma \quad\leq\quad f \quad\leq\quad \mu\!+\!\!B\sigma \quad\leq\quad f\!+\!\!2B\sigma.
\end{equation}
Since, we also know that $\sigma(x) \leq \epsilon_0$ for all $x \in \mc R_0$, we can conclude that
\[ f\!-\!\!2B\epsilon_0 \quad\leq\quad \mu\!-\!\!B\sigma \quad\leq\quad \mu\!+\!\!B\sigma \quad\leq\quad f\!+\!\!2B\epsilon_0. \]

Moreover, if condition $(\dagger)$ holds, we know that in $\mc R_0$, the function $f$ satisfies $-c_1\bs r^2 < f(x)-f(x_M) < -c_2\bs r^2$, where $\bs r = \bs r(x) := \|x-x_M\|$, so we get that  
\[ f(\!x_{\!M}\!)\!-\!c_1\!\bs r^2\!\!-\!\!2B\epsilon_0 \quad\leq\quad \mu\!-\!\!B\sigma \quad\leq\quad \mu\!+\!\!B\sigma \quad\leq\quad f(\!x_{\!M}\!)\!-\!c_2\bs r^2\!\!+\!\!2B\epsilon_0. \]
Now, recall that $\mc R_0$ is defined to consist of points $x$ where $\mu(\!x\!)\!+\!B\sigma(\!x\!) \geq \displaystyle\sup_{\mc D} \mu(\!x\!)\!-\!B\sigma(\!x\!)$, but given the fact that we have the above outer envelope for $\mu\pm B\sigma$, we can conclude that 
\begin{align*}
 \mc R_0 & \subseteq \left\{ x \,\Big|\, f(\!x_{\!M}\!)\!-\!c_2\bs r(x)^2\!\!+\!\!2B\epsilon_0 \;\geq\; \max f(\!x_{\!M}\!)\!-\!c_1\!\bs r(x)^2\!\!-\!\!2B\epsilon_0 \right\} \\
    & = \left\{ x \,\Big|\, f(\!x_{\!M}\!)\!-\!c_2\bs r(x)^2\!\!+\!\!2B\epsilon_0 \;\geq\; f(\!x_{\!M}\!)\!-\!2B\epsilon_0 \right\} \\
    & = \left\{ x \,\Big|\, -\!c_2\bs r(x)^2\!\!+\!\!2B\epsilon_0 \;\geq\; -\!2B\epsilon_0 \right\} \\
    & = \left\{ x \,\Big|\, c_2\bs r(x)^2 \;\leq\; 4B\epsilon_0 \right\} \\
    & = \left\{ x \,\Big|\, \bs r(x) \;\leq\; \sqrt{\frac{4B\epsilon_0}{c_2}} \right\}
\end{align*}

Now, if, on the other hand, $f$ satisfies condition $(\ddagger)$, then by the smoothness assumptions in $(\ddagger)$, we know that $\nabla f(x_M)$ is perpendicular to $\partial\mc D$ at $x_M$ and so there exist positive numbers $c_1$ and $c_2$ such that in a neighbourhood of $x_M$ we have
\[ -c_1\bs r \quad\leq\quad f-f(x_M) \quad\leq\quad -c_2\bs r^2. \]
Note that in the argument above in the case of $(\dagger)$, the precise form of the lower bound on $f$ was irrelevant, since all we are interested in is its maximum. So, the same argument goes through again.

This is depicted in Figure \ref{fig:PfBnd}, where $B := \sqrt{\beta_{N_0}} = \sqrt{b\ln N_0}$. 

\begin{figure*}[tb]
\begin{center}
  \includegraphics[width=\textwidth]{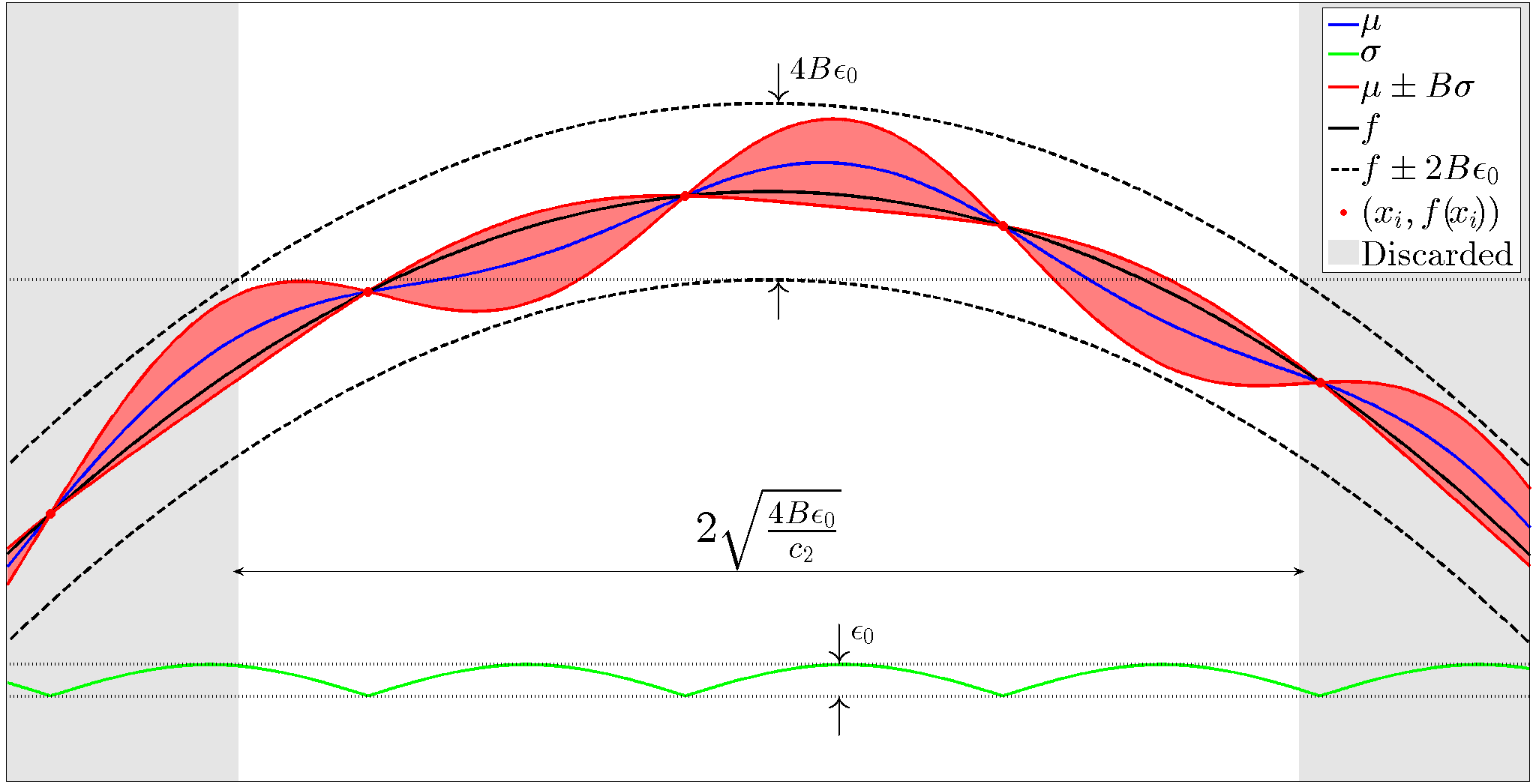}
\end{center}
\caption{\capstyle{The shrinking of the relevant set $\mc R_\ell$. Here, $B = \sqrt{\beta_{N_0}}$}} 
\label{fig:PfBnd}
\end{figure*}

\item[L$_{\ell+1}$:] Now, let us suppose that we are the end of the $\ell^{th}$ iteration. We have
\begin{align*}
N_{\ell+1} & \leq N_\ell + \mc N(\rho_\ell, \delta_\ell) \\
     & = N_\ell + \mc N\left(c\sqrt{\epsilon_\ell} \sqrt[4]{\ln N_\ell}, \delta_\ell\right) \\
     & \leq N_\ell + \mc N\left(c\sqrt{\frac{\epsilon_0}{4^\ell}} \sqrt[4]{\ln N_\ell}, \frac{\delta_0}{2^\ell}\right)  \qquad \text{by Proposition \ref{lem:varbound}} \\
     & = N_\ell + \mc N\left(c\sqrt{\epsilon_0} \sqrt[4]{\ln N_\ell}, \delta_0\right)  \qquad\quad \text{since $\mc N(2\rho,2\delta) = \mc N(\rho,\delta)$ for any $\rho$ and $\delta$} \\
     & \leq N_\ell + C(\ln N_\ell)^{\frac{d}{4}}
\end{align*}
So, the number of samples needed by the branch and bound algorithm is governed by the difference inequation
\begin{equation}
\Delta N_\ell \leq C(\ln N_\ell)^{\frac{d}{4}}.  \label{eqn:DiffEqn}
\end{equation}
To study the solutions of this difference equation, we consider the corresponding differential equation:
\begin{equation}
\frac{dN}{d\ell} = C(\ln N)^{\frac{d}{4}}.
\end{equation}
Since this equation is separable, we can write
\[ \frac{dN}{(\ln N)^{\frac{d}{4}}} = Cd\ell. \]
Now, letting $\ell = L$ be a given number of iterations in the algorithm and $N(L)$ the corresponding number of sampled points, we can integrate both sides of the above equation to get
\[ \int_{N(0)}^{N(L)} \frac{dN}{(\ln N)^{\frac{d}{4}}} = \int_0^L Cd\ell = CL. \]
Given the fact that the integral on the left can't be solved analytically, we will use the lower bound
\[ \frac{N(L)-N(0)}{(\ln N(L))^{\frac{d}{4}}} \leq \int_{N(0)}^{N(L)} \frac{dN}{(\ln N)^{\frac{d}{4}}} \]
to get 
\begin{equation}\label{eqn:ODESolUB} \frac{N(L)-N(0)}{C(\ln N(L))^{\frac{d}{4}}} \leq L  \end{equation}

Given a time $t$, we will denote by $\ell_t$ the largest non-negative integer such that $N_{\ell_t} < t$ or $0$ if no such number exists. We illustrate this somewhat obtuse definition with the following example:
\[ \xymatrix@=+2pt{%
\bullet & \bullet & \cdots & \bullet & \cdots & \bullet & \qquad\cdots\qquad & \bullet & \!\!\cdots\!\! & \bullet & \!\!\cdots\!\! & \!\!\bullet & \cdots \\
& & & & & & & & & & & & \\
1 \ar[uu] & 2 \ar[uu] &        & N_0 \ar[uu] &      & N_1 \ar[uu] &                  & N_{\ell_t}\!\! \ar[uu] & & t \ar[uu] &      & \!\!\!\!\!\!N_{\ell_t+1} \ar[uu] & } \]

Now, by Lemma \ref{lem:SKKS5.2}, for all $t >> N_0$ we have
\begin{align*}
r_t & \leq 2\sqrt{\beta_t} \max_{\mc R_{\ell_t}} \sigma_t \leq 2\sqrt{b \ln t} \epsilon_{\ell_t} \leq \frac{2\epsilon_0\sqrt{b \ln t}}{4^{\ell_t}} \leq \frac{8\epsilon_0\sqrt{b \ln t}}{4^{\ell_t+1}} \\
    & \leq 8\epsilon_0\sqrt{b\ln t}\left(\frac{1}{4}\right)^{\frac{N_{\ell_t+1}-N_0}{C\left(\ln N_{\ell_t+1}\right)^{d/4}}} \qquad \text{by Equation \ref{eqn:ODESolUB}} \\
    & \leq 8\epsilon_0\sqrt{b\ln t}\left(\frac{1}{4}\right)^{\frac{DN_{\ell_t+1}}{\left(\ln N_{\ell_t+1}\right)^{d/4}}} \qquad \text{for some $D > 0$ since $N_{\ell_t+1} > N_0$} \\
    & \leq 8\epsilon_0\sqrt{b\ln t}\left(\frac{1}{4}\right)^{\frac{Dt}{\left(\ln t\right)^{d/4}}} \qquad \text{for $t$ satisfying $\ln t > \frac{d}{4}$ (see $\star$ below) since } t \leq N_{\ell_t+1} \\
    & \leq 8\epsilon_0\sqrt{b}e^{-\frac{Et}{\left(\ln t\right)^{d/4}}+\frac{\ln\ln t}{2}} \\
    & \leq 8\epsilon_0\sqrt{b}e^{-\frac{Et}{\left(\ln t\right)^{d/4}}+\frac{Et}{2\left(\ln t\right)^{d/4}}} \qquad \text{for large enough $t$} \\
    & = Ae^{-\frac{\tau t}{\left(\ln t\right)^{d/4}}} \qquad \text{for $A = 8\epsilon_0\sqrt{b}$ and $\tau = E/2$.}
\end{align*}

\begin{itemize}
\item[$\star$] The reason for the specific criterion $\ln t > \frac{d}{4}$ is that the function $\frac{x}{(\ln x)^{d/4}}$ is increasing when this condition is satisfied, and so decreasing $x$ from $N_{\ell_t}+1$ to $t$ decreases its value, increasing the overall expression $\left( \frac{1}{4} \right)^{\frac{x}{(\ln x)^{d/4}}}$. To see that $\frac{x}{(\ln x)^{d/4}}$ becomes increasing when $\ln x > \frac{d}{4}$, we simply need to calculate its derivative:
\begin{align*}
   \frac{d}{dx} \frac{x}{(\ln x)^{d/4}} & = \frac{1}{(\ln x)^{d/4}} - \frac{d}{4}\frac{x}{x(\ln x)^{d/4+1}} \\
                     & = \frac{\ln x - \frac{d}{4}}{(\ln x)^{d/4}}.
\end{align*}
Moreover, since $N_{\ell_t+1} \geq t$, if the derivative of $\frac{x}{(\ln x)^{d/4}}$ is positive at $t$, it is also positive between $t$ and $N_{\ell_t+1}$ and so the function is indeed increasing in that interval.
\end{itemize}

\end{itemize}
\end{itemize}
\end{proof}

\end{document}